\pdfoutput=1

\documentclass[11pt]{article}

\usepackage{acl}

\usepackage{times}
\usepackage{latexsym}

\usepackage[T1]{fontenc}

\usepackage[utf8]{inputenc}

\usepackage{microtype}

\usepackage{inconsolata}

\usepackage{graphicx}
\usepackage{multirow}

\usepackage{amsmath}

\usepackage{relsize}

\usepackage{siunitx}

\usepackage{booktabs}
\usepackage{tabularx}
\usepackage{enumitem}

\usepackage{todonotes}

\newcommand{\biarrow}{$\leftrightarrow$}

\title{Machine Translation Models are\\ Zero-Shot Detectors of Translation Direction}

\author{Michelle Wastl \quad Jannis Vamvas \quad Rico Sennrich
\vspace{0.1cm}\\
 Department of Computational Linguistics, University of Zurich\\
\texttt{\{wastl,vamvas,sennrich\}@cl.uzh.ch}
}

\begin{document}
\maketitle
\begin{abstract}
Detecting the translation direction of parallel text is useful not only for machine translation training and evaluation but also has forensic applications, such as resolving plagiarism or forgery allegations.
In this work, we explore an unsupervised approach to translation direction detection based on the simple hypothesis that 
$p(\text{translation}|\text{original})>p(\text{original}|\text{translation})$, motivated by the well-known simplification effect in translationese or machine-translationese.
In experiments with multilingual machine translation models across 20 translation directions, we confirm the effectiveness of the approach for high-resource language pairs, achieving document-level accuracies of 82–96\% for NMT-produced translations, and 60–81\% for human translations, depending on the model used.\footnote{Code and demo are available at \url{https://github.com/ZurichNLP/translation-direction-detection}}
\end{abstract}

\section{Introduction}
\label{sec:intro}

While the original translation direction of parallel text is often ignored or unknown in the machine translation community, research has shown that it can be relevant for training~\cite{kurokawa-etal-2009-automatic,ni-etal-2022-original} and evaluation~\cite{graham-etal-2020-statistical}.\footnote{As of today, training data is not typically filtered by translation direction, but we find evidence of a need for better detection in recent work. For example, \citet{post2023escaping} show that back-translated data is more suited than crawled parallel data for document-level training, presumably because of translations in the crawled data that lack document-level consistency.}
Beyond machine translation, translation direction detection has practical applications in areas such as forensic linguistics, where determining the original of a document pair may help resolve plagiarism or forgery accusations.

Previous work has addressed translation (direction) detection with feature-based approaches, using features such as n-gram frequency statistics and POS tags for classification \cite{kurokawa-etal-2009-automatic,10.1093/llc/fqt031,Sominsky2019} or unsupervised clustering \cite{Nisioi2015, Rabinovich2015}.
However, these methods require a substantial amount of text data, and cross-domain differences in the statistics used can overshadow differences between original and translationese text.

\begin{figure}
    \centering
    \includegraphics[width=\columnwidth, trim=0 0.15cm 0 0, clip]{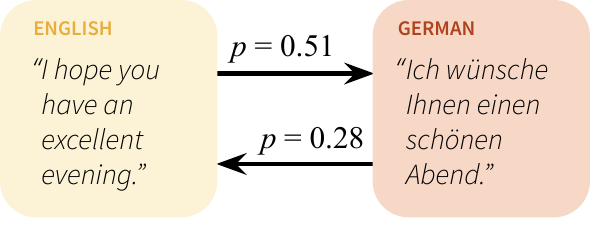}
    \caption{
        NMT models can be used for inferring the likely original translation direction of parallel text.
    In this example, the NMT model assigns a much higher probability to the German sentence given the English sentence than to the English sentence given the German sentence, indicating that the more likely original translation direction is English\(\rightarrow\)German.
    }
    \label{fig:figure1}
\end{figure}

\begin{figure*}
    \begin{center}
      \includegraphics[width=0.85\textwidth]{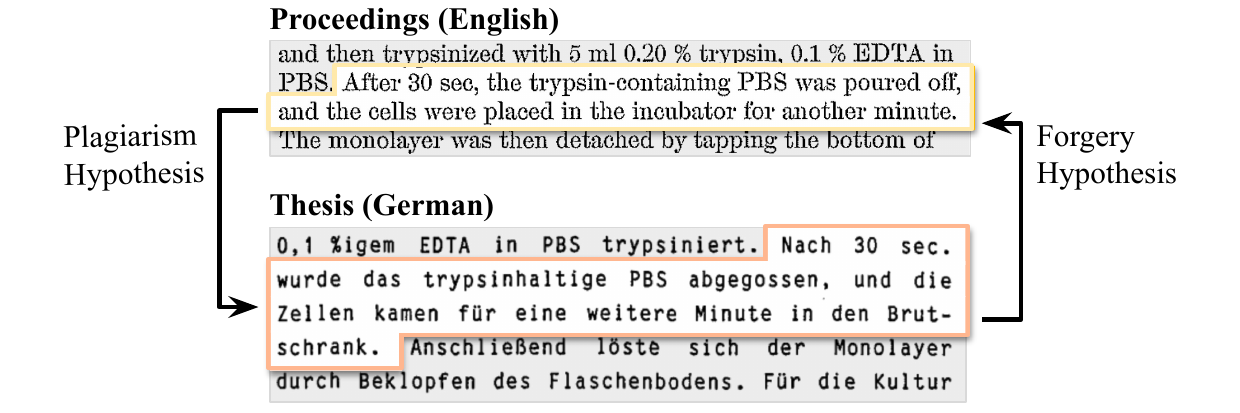}
    \end{center}
    \caption{A recent forensic case in Germany underscores the relevance of translation direction detection~\cite{ebbinghaus2022b, zenthoefer2022b, dewiki:238411824}.
    In 2022, two experts raised concerns about the originality of a German PhD thesis and suspected it to be plagiarized from a proceedings volume in English (\textit{plagiarism hypothesis}).
    Further investigation showed, however, that the alleged English source could not be found in any library or database. This raised the possibility of a deliberate attempt to discredit the thesis author by fabricating the English book (\textit{forgery hypothesis}).
    Initially, the debate focused on the dating of the typefaces and paper used to print the book, in addition to textual inconsistencies.
    A computational analysis of translation direction could provide additional evidence in this or similar cases.
    The illustration depicts one of the parallel passages identified by \citet{Weber2022}.
    }
    \label{fig:figure2}
  \end{figure*}

In this work, we explore the unsupervised detection of translation directions purely on the basis of a multilingual neural machine translation (NMT) system's translation probabilities.
As illustrated in Figure~\ref{fig:figure1}, we hypothesize that $p(\text{translation}|\text{original})>p(\text{original}|\text{translation})$, which, if it generally holds, would allow us to infer the original translation direction.

If the translation has been automatically generated, this hypothesis can be motivated by the fact that machine translation systems typically generate text with mode-seeking search algorithms, and consequently tend to over-produce high-frequency outputs and reduce lexical diversity \cite{vanmassenhove-etal-2019-lost}.
However, even human translations are known for so-called translationese properties such as interference, normalization, and simplification, and a (relative) lack of lexical diversity \cite{Teich+2003, 10.1093/llc/fqt031,toral-2019-post}.

We test the approach on 20 translation directions, experimenting with 3 massively multilingual NMT models to predict the translation probabilities of human translations, NMT-produced translations, LLM-generated translations, and pre-neural machine translations.
We find that the approach detects the translation direction of human translations with an average accuracy of 66\% at the sentence level, and 80\% for documents with $\geq$ 10 sentences.
For the output of NMT systems, detection accuracy is even higher, but our hypothesis that $p(\text{translation}|\text{original})>p(\text{original}|\text{translation})$ does not hold for the output of pre-neural systems.

To compare our unsupervised approach to a supervised baseline, we implement a modernized version of the approach proposed by \citet{Sominsky2019}.
A controlled comparison shows that a supervised approach can outperform our unsupervised one under ideal conditions (with in-domain labelled training data) for human translations, but performance drops in a cross-domain setting. Notably, for NMT-produced translations, the unsupervised approach remains competitive even when tested within the same domain.

Finally, we apply our method to a recent forensic case~(Figure~\ref{fig:figure2}), where the translation direction of a German PhD thesis and an English book has been under dispute, finding additional evidence for the hypothesis that the English book is a forgery created to make the thesis appear plagiarized.

\medskip

\noindent{}Our main contributions are the following:
\begin{itemize}[itemsep=0pt]
    \item We propose a simple, unsupervised approach to translation direction detection based on the translation probabilities of NMT models.
    \item We demonstrate that the approach is effective for detecting the original translation direction of neural machine translations, and to a lesser extent, human translations in a variety of high-resource language pairs.
    \item We provide a qualitative analysis of detection performance and apply the method to a real-world forensic case.
\end{itemize}

\section{Related Work}
\label{sec:rel_work}

\subsection{Translation (Direction) Detection}
\label{subsec:td}

In an ideal scenario where large-scale annotated in-domain data is available, high accuracy can be achieved in translation direction detection at phrase and sentence level by training supervised systems based on various features such as word frequency statistics, POS n-grams or text embeddings \cite{Sominsky2019}.

To reduce reliance on in-domain supervision, unsupervised methods that rely on clustering and consequent cluster labelling have also been explored for the related task of translationese detection~\cite{Rabinovich2015, Nisioi2015}.
One could conceivably perform translation direction detection using similar methods, but this has the practical problem of requiring an expert for cluster labelling and poor open-domain performance.
In a multi-domain scenario, \citet{Rabinovich2015} observe that clustering based on features proposed by \citet{10.1093/llc/fqt031} results in clusters separated by domain rather than translation status.
They address this by producing $2k$ clusters, $k$ being the number of domains in their dataset, and labelling each.
Clearly, labelling becomes more costly as the number of domains increases, which limits applicability to an open-domain scenario.

In contrast, we hypothesize that comparing translation probabilities remains a valid strategy across domains, and requires no resources other than NMT models that are competent for the respective language pair.

\subsection{Translation Probabilities}
\label{subsec:tb}

Previous work has leveraged translation probabilities for tasks such as noisy parallel corpus filtering \cite{Junczys-Dowmunt2018}, machine translation evaluation \cite{Thompson2020}, and paraphrase identification \cite{mallinson-etal-2017-paraphrasing,vamvas-sennrich-2022-nmtscore}. 
Those approaches analyze translation probabilities symmetrically in two directions, which is also the case in this work.

\section{Methods}
\label{sec:methods}

Given a parallel sentence pair $(x,y)$, the main task in this work is to identify the translation direction between a language X and a language Y, and, consequently, establish which side is the original and which is the translation.
This is achieved by comparing the conditional translation probability $P(y|x)$ by an NMT model $M_{X\rightarrow Y}$ with the conditional translation probability $P(x|y)$ by a model $M_{Y\rightarrow X}$ operating in the inverse direction.
Our core assumption is that segment pairs in the original translation direction are assigned higher conditional probabilities by NMT models than in the inverse direction, so if $P(y|x) > P(x|y)$, we predict that $y$ is the translation, and $x$ the original, and the original translation direction is $X \to Y$.

\subsection{Detection on the Sentence Level}

With a probabilistic autoregressive NMT model, we can obtain $P(y|x)$ as a product of the individual token probabilities:
\begin{equation}
    \label{eq:avg}
    P(y|x) = \prod_{j=1}^{|y|} p(y_j|y_{<j},x)
\end{equation}
\vspace{0.2cm}

\noindent We follow earlier work by \citet{Junczys-Dowmunt2018,Thompson2020}, and average token-level (log-)probabilities.\footnote{
The models we use have been trained with label smoothing~\cite{szegedy2016rethinking}, which has a cumulative effect on sequence-level probabilities~\cite{yan2023dcmbr}. Averaging token-level probabilities can help mitigate this shortcoming.}

\begin{equation}
    \label{eq:avg2}
    P_{\text{tok}}(y|x) = P(y|x)^{\frac{1}{|y|}}
\end{equation}
To detect the original translation direction (OTD), $P_{\text{tok}}(y|x)$ and $P_{\text{tok}}(x|y)$ are compared:

\[
\begin{aligned}
\text{OTD} = 
\begin{cases} 
  X \to Y, & \text{if } P_{\text{tok}}(y|x) > P_{\text{tok}}(x|y) \\
  Y \to X, & \text{otherwise}
\end{cases}
\end{aligned}
\]

\subsection{Detection on the Document Level}\label{subsec:doc_level}
We also study translation direction detection at the level of documents, as opposed to individual sentences.
We assume that the sentences in the document are aligned 1:1, so that we can apply an NMT model trained at the sentence level to all $n$~sentence pairs $(x_i, y_i)$ in the document, and then aggregate the result.

Our approach is equivalent to the sentence-level approach in that we calculate the average token-level probability across the document, conditioned on the respective sentence in the other language:
\begin{equation}
    \label{eq:avg_doc}
    P_{\text{tok}}(y|x) = [\prod_{i=1}^{n} \prod_{j=1}^{|y_i|} p(y_{i,j}|y_{i,<j},x_i)]^{\frac{1}{\scriptstyle{|y_1| + \dots + |y_n|}}}
\end{equation}
The criterion for the original translation direction is then again whether $P_{\text{tok}}(y|x) > P_{\text{tok}}(x|y)$.

\subsection{On Directional Bias}
\label{subsec:bn}

A multilingual translation model (or a pair of bilingual models) may consistently assign higher probabilities in one translation direction than the other, thus biasing our prediction.
This could be the result of training data imbalance, tokenization choices, or typological differences between the languages~\cite{cotterell-etal-2018-languages,bugliarello-etal-2020-easier}.\footnote{We note that \citet{bugliarello-etal-2020-easier} do not control for the original translation direction of their data. Re-examining their findings in view of our core hypothesis could be fruitful future work.}

To allow for a cross-lingual comparison of bias despite varying data balance, we measure bias via the difference in accuracy between the two gold directions.
An unbiased model should have similar accuracy in both.
An extremely biased model that always predicts $\text{OTD}=X \to Y$ will achieve perfect accuracy on the gold direction $X \to Y$, and zero accuracy on the reverse gold direction $Y \to X$.
We will report the bias $B$ as follows:
\begin{equation}
    \label{eq:avg_doc}
B=|acc(X \rightarrow Y)-acc(Y \rightarrow X)|
\end{equation}
This yields a score that ranges from 0 (unbiased) to~1 (fully biased).

\section{Experiments: Models and Data}
\label{subsec:models}

\subsection{Unsupervised Setting}

We experiment with three multilingual machine translation models:
M2M-100-418M \citep{Fan2021}, SMaLL-100 \citep{Mohammadshahi2022}, and NLLB-200-1.3B \citep{nllb}.

The models are architecturally similar, all being based on the Transformer architecture \cite{DBLP:journals/corr/VaswaniSPUJGKP17}, but they differ in the training data used, number of languages covered, model size, and, consequently, in translation quality.
The comparison allows conclusions about how sensitive our method is to translation quality -- NLLB-200-1.3B yields the highest translation quality of the three~\citep{tiedemann-de-gibert-2023-opus}, but we also highlight differences in data balance.
English has traditionally been dominant in the amount of training data, and all three models aim to reduce this dominance in different ways, for example via large-scale back-translation \cite{sennrich-etal-2016-improving} in M2M-100-418M and NLLB-200-1.3B.
SMaLL-100 is a distilled version of the M2M-100-12B model, and samples training data uniformly across language pairs.

\begin{table}[b!]
\smaller[2]
\centering
\begin{tabular}{lrrrrr}
\toprule
& \multicolumn{2}{c}{Source} & \multicolumn{3}{c}{Target Sentences} \\
Direction & Sents & Docs ($\geq$10) & HT & NMT & Pre-NMT \\
\cmidrule(r){2-3} \cmidrule(l){4-6}
cs$\rightarrow$en & 1448 & 129 & 2896 & 15928 & 16489 \\
cs$\rightarrow$uk & 3947 & 112 & 3947 & 49381 & - \\
de$\rightarrow$en & 1984 & 121 & 3968 & 17856 & 13491 \\
de$\rightarrow$fr & 1984 & 73 & 1984 & 11904 & - \\
en$\rightarrow$cs & 4111 & 204 & 6148 & 51480 & 27000 \\
en$\rightarrow$de & 2037 & 125 & 4074 & 18333 & 18000 \\
en$\rightarrow$ru & 4111 & 174 & 4111 & 47295 & 15000 \\
en$\rightarrow$uk & 4111 & 174 & 4111 & 41147 & - \\
en$\rightarrow$zh & 4111 & 204 & 6148 & 57591 & - \\
fr$\rightarrow$de & 2006 & 71 & 2006 & 14042 & - \\
ru$\rightarrow$en & 3739 & 136 & 3739 & 40836 & 13482 \\
uk$\rightarrow$cs & 2812 & 43 & 2812 & 33744 & - \\
uk$\rightarrow$en & 3844 & 88 & 3844 & 40266 & - \\
zh$\rightarrow$en & 3851 & 162 & 5726 & 52140 & - \\
\addlinespace
Total & 44096 & 1816 & 55514 & 491943 & 103462 \\
\bottomrule
\end{tabular}
\caption{Statistics of the WMT data used in our main experiments. More granular statistics are provided in Appendix~\ref{sec:appendix_data}.}
\label{tab:hr_stats}
\end{table}

We test the approach on datasets from the WMT news/general translation tasks from WMT16~\cite{bojar-etal-2016-findings}, WMT22~\cite{kocmi-etal-2022-findings}, and WMT23~\cite{kocmi-etal-2023-findings}, which come annotated with document boundaries and the original language of each document. We include different years of WMT data not only to test the approach for different translation directions and translation types but also to rule out test set contamination for experiments with models, for which the test set predates the model release. The results for the individual WMT datasets are reported in Appendix~\ref{app_contamination}.

We also experiment with a subset of the FLORES-101 dataset~\cite{goyal-etal-2022-flores} to test the approach on indirect translations, where English was the original language of both sides of the parallel text.
We divide the data into subsets based on several categorizations:
\vspace{0.2cm}

\begin{itemize}
\item \textbf{Translation direction}: the WMT data span 14 translation directions and 3 scripts (Latin, Cyrillic, Chinese).
\item \textbf{Type of translation}: we distinguish between human translations (\textbf{HT}), which consist of (possibly multiple) reference translations, neural translation systems (\textbf{NMT}) (WMT 2016; 2022–2023; (incl. translations from machine translation systems specifically and, to a lesser extent, translations produced by large language models (LLMs)), and phrase-based or rule-based pre-neural systems from WMT 2016 (\textbf{pre-NMT}) as a third category.
\item \textbf{Directness}: Given that the WMT data are \textit{direct} translations from one side of the parallel text to the other, we perform additional experiments on translations for~6 \mbox{FLORES} language pairs (Bengali\biarrow\allowbreak Hindi, Czech\biarrow\allowbreak Ukrainian, German\biarrow\allowbreak French, German\biarrow\allowbreak Hindi, Chinese\biarrow\allowbreak French, and Xhosa\biarrow\allowbreak Zulu). This allows us to analyze the behavior of our approach on \textit{indirect} sentence pairs where both sequences are translations from a third language (in this case, English).

\end{itemize}

\begin{table*}[t!]
\centering
\begin{tabularx}{\textwidth}{lXrrrrrr}
\toprule
& & \multicolumn{3}{c}{HT} & \multicolumn{3}{c}{NMT} \\
\cmidrule(lr){3-5} \cmidrule(lr){6-8}
LP & & \(\rightarrow\) & \(\leftarrow\) & Avg. & \(\rightarrow\) & \(\leftarrow\) & Avg. \\
\midrule
\multirow{2}{*}{en\biarrow cs}
& supervised   & 64.70 & \textbf{71.36} & \textbf{68.03} & 65.43 & 71.98 & 68.71 \\
& unsupervised & \textbf{68.85} & 65.19 & 67.02 & \textbf{71.87} & \textbf{78.30} & \textbf{75.09} \\
\addlinespace
\multirow{2}{*}{en\biarrow ru}
& supervised   & 56.65 & \textbf{81.74} & \textbf{69.19} & 61.72 & \textbf{82.97} & 72.35 \\
& unsupervised & \textbf{71.81} & 54.05 & 62.93 & \textbf{76.91} & 71.98 & \textbf{74.44} \\
\addlinespace
\multirow{2}{*}{en\biarrow de}
& supervised   & \textbf{87.05} & 55.29 & \textbf{71.17} & \textbf{89.52} & 61.08 & \textbf{75.30} \\
& unsupervised & 56.38 & \textbf{67.44} & 61.91 & 62.69 & \textbf{85.27} & 73.98 \\
\addlinespace
\multirow{2}{*}{Macro Avg.}
& supervised   & \textbf{69.46} & \textbf{69.47} & \textbf{69.46} & \textbf{72.22} & 72.01 & 72.12 \\
& unsupervised & 65.68 & 62.23 & 63.95 & 70.49 & \textbf{78.51} & \textbf{74.50} \\
\bottomrule
\end{tabularx}
\caption{Accuracy of the proposed unsupervised approach, using the M2M100 MT system, compared to the performance of XLM-R fine-tuned on the translation direction detection task with WMT data.}
\label{tab:supervised_results}
\end{table*}

We use HT and NMT translations from WMT16 as a validation set and the remaining translations for testing our approach.
Table~\ref{tab:hr_stats} shows test set statistics for our main experiments.

To demonstrate our approach on ``real-world'' data, we use text pertaining to a publicly documented plagiarism allegation case, in which translation-based plagiarism was the main focus~\citep{ebbinghaus2022b, zenthoefer2022b}.
We use 86 parallel segments in German and English from aligned excerpts of both the PhD thesis and the alleged source that were presented in a plagiarism analysis report \citep{Weber2022}. We extracted the segments with OCR and manually checked for OCR errors. An example translation from this dataset is given in Appendix~\ref{app_real_world}.

\subsection{Supervised Baseline}
\label{subsec:supervised}
In addition to the unsupervised approach, we fine-tune XLM-R (base)~\cite{conneau-etal-2020-unsupervised} on the translation direction detection task to provide a supervised baseline for our experiments. 

We extract the final hidden state for the first token by XLM-R for each segment of a translation, which serves as the feature representation for that segment. Inspired by the COMET architecture for MT evaluation~\cite{Rei2020}, we then combine the resulting representations for each translation by concatenating the addition of the representations of the segments of a pair, their absolute difference, and their product.\footnote{We deviate slightly from COMET by using the addition of the two segment representations instead of concatenating them sequentially, to avoid introducing input order bias.}$^,$\footnote{Furthermore, we experiment with encoding the source segment jointly with the translation. This yields similar results, but introduces input order bias. The test set results for this alternative architecture are listed in Appendix~\ref{app:joint_enc}.}
The resulting concatenation is used as input to train the classification head.  

For the training set, we take a sample of 1398–1400 source segments and their corresponding HT and NMT-based translations per direction from WMT16. Additionally, we sample another 100 source segments per direction and their corresponding HT, NMT and pre-NMT-based translations from WMT16 as a validation set. We then train bilingual classifiers for each en\biarrow cs, en\biarrow ru and en\biarrow de. We validate settings with different learning rates and epochs (Appendix~\ref{app_supervised}) and report the results for the test set in Table~\ref{tab:supervised_results} with the setting that achieves the highest accuracies for the corresponding language pair\footnote{en\biarrow cs/ru: learning rate 1e-05, epoch 2; en\biarrow de: learning rate 1e-05, epoch 4.} on the validation set. The test set consists of HT and NMT sentence pairs from WMT21, WMT22 and WMT23, which is equivalent to the unsupervised experiments apart from the WMT16.  

Since training the supervised systems on larger datasets could improve their performance, we experiment with training them on subsets of the Europarl corpus~\cite{Koehn2005, Ustaszewski}. Although higher accuracies are reached when the systems are tested on in-domain data, their performance drops below the WMT-based system when tested on out-of-domain data (WMT).  Hence, we choose to describe the WMT-based model, which performs best on WMT data in the main part of this paper (Subsection~\ref{subsec:hr}) and give a detailed description and results of the Europarl-based system in Appendix~\ref{app_europarl}.

\section{Results}
\label{sec:results}

\subsection{Sentence-level Classification}
\label{subsec:hr}

\begin{table*}[h!]
\centering
\begin{tabularx}{\textwidth}{@{}Xrrrrrrrrr@{}}
\toprule
& \multicolumn{3}{c}{M2M-100-418M} & \multicolumn{3}{c}{SMaLL-100} & \multicolumn{3}{c}{NLLB-200-1.3B} \\
\cmidrule(lr){2-4} \cmidrule(lr){5-7} \cmidrule(lr){8-10}
Language Pair & \(\rightarrow\) & \(\leftarrow\) & Avg. & \(\rightarrow\) & \(\leftarrow\) & Avg. & \(\rightarrow\) & \(\leftarrow\) & Avg. \\
\midrule
HT~~en\biarrow cs & 68.85 & 65.19 & \textbf{67.02} & 63.08 & 69.37 & 66.22 & 54.05 & 68.78 & 61.42 \\
HT~~en\biarrow de & 56.38 & 67.44 & \textbf{61.91} & 58.62 & 63.10 & 60.86 & 59.70 & 47.76 & 53.73 \\
HT~~en\biarrow ru & 71.81 & 54.05 & 62.93 & 68.38 & 57.56 & \textbf{62.97} & 67.40 & 49.08 & 58.24 \\
HT~~en\biarrow uk & 71.95 & 69.56 & \textbf{70.76} & 70.49 & 68.83 & 69.66 & 47.21 & 64.00 & 55.61 \\
HT~~en\biarrow zh & 54.25 & 84.30 & \textbf{69.27} & 56.41 & 80.54 & 68.48 & 17.81 & 82.52 & 50.16 \\
HT~~cs\biarrow uk & 52.44 & 74.40 & 63.42 & 59.26 & 70.52 & \textbf{64.89} & 47.68 & 76.67 & 62.18 \\
HT~~de\biarrow fr & 89.72 & 50.50 & 70.11 & 85.48 & 57.68 & 71.58 & 86.29 & 62.16 & \textbf{74.23} \\
\addlinespace
Macro-Avg. & 66.49 & 66.49 & \textbf{66.49} & 65.96 & 66.80 & 66.38 & 54.31 & 64.42 & 59.37 \\
\bottomrule
\end{tabularx}
\caption{Accuracy of three different models when detecting the direction for human translation.
The first column per model reports accuracy for sentence pairs with left-to-right gold direction~(e.g., en\(\rightarrow\)cs), the second column for sentence pairs with the reverse gold direction~(e.g., en\(\leftarrow\)cs). The last column reports the macro-average across both directions.
The best average result for each language pair is printed in bold.
}
\label{tab:full_results_ht}
\end{table*}

\begin{table}
\centering
\begin{tabularx}{\columnwidth}{@{}Xrrr@{}}
\toprule
Language Pair &  \(\rightarrow\) &  \(\leftarrow\) & Avg. \\
\midrule
NMT~~en\biarrow cs & 71.87 & 78.30 & 75.09 \\
NMT~~en\biarrow de & 62.69 & 85.27 & 73.98 \\
NMT~~en\biarrow ru & 76.91 & 71.98 & 74.44 \\
NMT~~en\biarrow uk & 75.01 & 79.31 & 77.16 \\
NMT~~en\biarrow zh & 64.29 & 90.29 & 77.29 \\
NMT~~cs\biarrow uk & 72.83 & 79.15 & 75.99 \\
NMT~~de\biarrow fr & 90.65 & 51.60 & 71.13 \\
\addlinespace
Macro-Avg. & 73.46 & 76.56 & 75.01 \\
\bottomrule
\end{tabularx}
\caption{Accuracy of M2M-100 when detecting the translation direction of NMT-produced translations.}
\label{tab:hr_results_nmt}
\end{table}

\begin{table}
\centering
\begin{tabularx}{\columnwidth}{@{}Xrrr@{}}
\toprule
Language Pair &  \(\rightarrow\) &  \(\leftarrow\) & Avg. \\
\midrule
Pre-NMT~~en\biarrow cs & 41.97 & 42.59 & 42.28 \\
Pre-NMT~~en\biarrow de & 33.33 & 54.30 & 43.81 \\
Pre-NMT~~en\biarrow ru & 37.98 & 39.01 & 38.49 \\
\addlinespace
Macro-Avg. & 37.76 & 45.30 & 41.53 \\
\bottomrule
\end{tabularx}
\caption{Accuracy of M2M-100 when detecting the translation direction of sentences translated with \mbox{pre-NMT} systems.}
\label{tab:hr_results_pnmt}
\end{table}

\begin{table}
\centering
\begin{tabularx}{\columnwidth}{@{}Xrrr@{}}

\toprule
Language Pair &  \(\rightarrow\) &  \(\leftarrow\) & Avg. \\
\midrule
en\biarrow ru & 75.75 & 68.08 & 71.91 \\
en\biarrow uk & 72.76 & 72.56 & 72.66 \\
en\biarrow zh & 58.82 & 90.54 & 74.68 \\
\addlinespace
Macro-Avg. & 69.11 & 77.06 & 73.08 \\
\bottomrule
\end{tabularx}
\caption{Accuracy on translations generated by LLMs with M2M-100.}
\label{tab:results_llm}
\end{table}

The sentence-level results are shown in Tables~\ref{tab:full_results_ht}~(HT),~\ref{tab:hr_results_nmt}~(NMT), and~\ref{tab:hr_results_pnmt}~(pre-NMT).

Table~\ref{tab:full_results_ht} compares the results for HT across all models. As a general result, we find that it is not NLLB, but M2M-100 that on average yields the best results for HT on the sentence level, with SMaLL-100 a close second. Hence, we report results for experiments with M2M-100 in the following, while performance of the other models is reported in Appendix~\ref{appA} and~\ref{appB}.

A second result is that the translation detection works best for NMT-based translations (75.0\% macro-average), second-best for HT (66.5\% macro-average), and worst for pre-NMT (41.5\% macro-average).
The fact that performance for pre-neural systems is below chance level indicates that the NMT systems we use tend to assign low probabilities to the (often ungrammatical) outputs of pre-neural systems.

A third result is that accuracy varies by language pair. Among the language pairs tested, accuracy of M2M-100 ranges from 61.9\% (en$\leftrightarrow$de) to 70.8\% (en$\leftrightarrow$uk) for HT, and from 71.1\% (de$\leftrightarrow$fr) to 77.3\% (en$\leftrightarrow$zh) for NMT.



In comparison to the results of the supervised system, we show in Table~\ref{tab:supervised_results} that while our unsupervised approach is outperformed on HT, it is competitive for NMT-based translation. Taking into consideration the (cross-domain) results of the Europarl-based models shown in Table~\ref{tab:europarl1} as well, the main benefits of the unsupervised approach are highlighted: independence from training data and flexibility across languages and domains.

\begin{table}
\centering
\begin{tabularx}{\columnwidth}{@{}Xrr@{}}
\toprule
Language Pair &  Ratio of \(\rightarrow\) &  Ratio of \(\leftarrow\)\\
\midrule
HT~~bn\biarrow hi & 66.80\,\% & 33.20\,\% \\
HT~~de\biarrow hi & 48.72\,\% & 51.28\,\% \\
HT~~cs\biarrow uk & 42.69\,\% & 57.31\,\% \\
HT~~de\biarrow fr & 84.78\,\% & 15.22\,\% \\
HT~~zh\biarrow fr & 91.70\,\% & 8.30\,\% \\
HT~~xh\biarrow zu & 45.06\,\% & 54.94\,\% \\
\bottomrule
\end{tabularx}
\caption{Percentage of predictions by M2M-100 for each translation direction when neither is the true translation direction (English-original FLORES).}
\label{tab:lr_results_ht}
\end{table}

\begin{table*}
\centering
\smaller
\begin{tabularx}{\textwidth}{@{}rXrrr@{}}
  \toprule
  & Sentences & \(\rightarrow\) & \(\leftarrow\) & Rel. Difference  \\
  \midrule
  1 & \textit{DE: Mit dem Programm "Guten Tag, liebes Glück" ist er seit 2020 auf Tour.} & & &  \\
  & EN: He has been on tour with the programme "Guten Tag, liebes Glück" since 2020. (HT) &  0.145 & 0.558 & \phantom{0}0.26 \\
  & EN: He has been on tour since 2020. (NMT) & 0.272 & 0.092 & \textbf{\phantom{0}2.95}  \\
  \addlinespace
  2 & \textit{EN: please try to perfprm thsi procedures"} & & &  \\
  & DE: bitte versuchen Sie es mit diesen Verfahren (HT) & 0.246 & 0.010 & \textbf{24.29} \\
  & DE: Bitte versuchen Sie, diese Prozeduren durchzuführen" (NMT) & 0.586 & 0.025 & \textbf{23.59}  \\
  \addlinespace
  3 & \textit{EN: If costs for your country are not listed, please contact us for a quote.} & & &  \\
  & DE: Wenn die Kosten für Ihr Land nicht aufgeführt sind, wenden Sie sich für einen Kostenvoranschlag an uns. (HT) & 0.405 & 0.525 & \phantom{0}0.77  \\
  & DE: Wenn die Kosten für Ihr Land nicht aufgeführt sind, kontaktieren Sie uns bitte für ein Angebot. (NMT) & 0.697 & 0.585 & \textbf{\phantom{0}1.19}\\
  \addlinespace
  4 & \textit{EN: Needless to say, it was chaos.} & & &  \\
  & DE: Es war natürlich ein Chaos. (HT) & 0.119 & 0.372 & \phantom{0}0.32 \\
  & DE: Unnötig zu sagen, es war Chaos. (NMT) & 0.755 & 0.591 & \textbf{\phantom{0}1.28}  \\
  \addlinespace
  5 & \textit{DE: Mit freundlichen Grüßen} & & &  \\
  & FR: Cordialement (HT) & 0.026 & 0.107 & \phantom{0}0.24  \\
  & FR: Sincèrement (NMT) & 0.015 & 0.083 & \phantom{0}0.18  \\
  & FR: Sincères amitiés (NMT) & 0.062 & 0.160 & \phantom{0}0.39  \\
  & FR: Avec mes meilleures salutations (NMT) & 0.215 & 0.353 & \phantom{0}0.61  \\
  \bottomrule
\end{tabularx}
\caption{Qualitative comparison of sentence pairs. Source sentences are marked in \textit{italics}, and gold direction is always \(\rightarrow\). Relative probability difference $>1$ indicates that translation direction was successfully identified, and is highlighted in bold.
The probabilities are generated by M2M-100.}
\label{tab:quali_analysis}
\end{table*}

\subsection{LLM-generated Translations}
The main focus of this paper lies on three different translation types: HT, NMT, and pre-NMT. However, LLMs have also shown strong translation capabilities~\citep{kocmi-etal-2023-findings}. Hence, GPT-4 was considered as one of the translation systems in the WMT23 shared task and its outputs are therefore part of the NMT test set in this work. Table~\ref{tab:results_llm} shows the results of our translation direction detection approach on the GPT-4-generated test subset in isolation to document its performance on this additional translation type. For all three language pairs, our approach reaches accuracies that are comparable to the ones reached for the same pairs' NMT translations. 

\subsection{Indirect Translations}
With an experiment on the English-original FLORES data, we evaluate our approach on the special case that neither side is the original.
As shown in Table~\ref{tab:lr_results_ht}, our approach yields relatively balanced predictions on human translations for cs\biarrow uk and xh\biarrow zu, predicting each direction a roughly equal number of times.
For de\biarrow fr, we again find that the model predicts de\(\rightarrow\)fr more frequently than the reverse direction. The result for fr\biarrow zh display an even larger degree of disparity.

\subsection{Directional Bias}
\label{subsec:bias}
When analyzing Table~\ref{tab:full_results_ht} for directional bias, we observe that M2M-100 is especially biased in the directions~de$\to$fr ($B=0.39$) and zh$\to$en ($B=0.30$).
While we expected a general bias towards x$\to$en due to the dominance of English in training data, we find that the direction and strength of the bias vary across language pairs and models.
An extreme result is NLLB for en\biarrow zh, with $B=0.64$ towards zh$\to$en. 

There appears to be an inclination for bias towards languages closely related to English for some language pairs, such as de\biarrow fr and zh\biarrow fr (Table~\ref{tab:lr_results_ht}) in contrast to language pairs, with both languages being closely related to (and including) English, such as en\biarrow fr. The discrepancy in balance between de\biarrow hi and zh\biarrow fr indicates, however, that relatedness to English might not have a strong general influence on our approach, but rather that there is a bias for $\rightarrow$French and Chinese$\rightarrow$ when applying it with M2M-100 or SMaLL-100.

Furthermore, the supervised systems exhibit higher bias (Table~\ref{tab:supervised_results}) than the unsupervised approach for all language pairs and translation types, indicating that directional bias is an issue for both supervised and unsupervised approaches. 

We leave it to future work to explore whether bias can be reduced via different normalization strategies, a language pair specific bias correction term, or different model training.
At present, our recommendation is to be mindful in the choice of NMT model and to perform validation before trusting the results of a previously untested NMT model for translation direction detection.

\subsection{Qualitative Analysis}
\label{subsec:ea}

A qualitative comparison of sources and translations, as illustrated in Table \ref{tab:quali_analysis}, reveals that factors such as normalization, simplification, word order interference, and sentence length influence the detection of translation direction.
In Example 1, an English HT translates the German source fully, while the NMT omits half of the content, showing a high degree of simplification. Our method recognizes the simplified NMT version as a translation, but not the more complete HT. 
Example 2 demonstrates normalization to a high degree. This shows that NMT-models assign extremely low probabilities when the target is more ungrammatical than the source.
The third example indicates that translations exhibiting normalization, simplification, and interference to a higher degree are more likely to be identified. 
In Example 4, source language interference in terms of word order and choice significantly impacts the detection; the more literal translation mirroring the source's word order is recognized, while the more liberal translation is not. 
Finally, Example 5 highlights challenges with short sentences: The German phrase \textit{Mit freundlichen Grüßen} is fairly standardized, while its French equivalents can vary in use and context, adding to the ambiguity and affecting the probability distribution in NMT. Hence, our approach fails to identify any of the French translations without additional context.

\begin{table}
\centering
\begin{tabularx}{\columnwidth}{@{}Xrrr@{}}
\toprule
Language Pair &  \(\rightarrow\) &  \(\leftarrow\) & Avg. \\
\midrule
HT~~en\biarrow cs & 88.24 & 80.62 & 84.43 \\
HT~~en\biarrow de & 70.40 & 88.43 & 79.41 \\
HT~~en\biarrow ru & 96.55 & 54.41 & 75.48 \\
HT~~en\biarrow zh & 67.65 & 97.53 & 82.59 \\
\addlinespace
Macro-Avg. & 80.71 & 80.25 & 80.48 \\
\bottomrule
\end{tabularx}
\caption{Document-level classification: Accuracy of M2M-100 when detecting the translation direction of human translations at the document level~(documents with $\geq$ 10 sentences).}
\label{tab:doc_ht}
\end{table}

\begin{table}
\centering
\begin{tabularx}{\columnwidth}{@{}Xrrr@{}}
\toprule
Language Pair &  \(\rightarrow\) &  \(\leftarrow\) & Avg. \\
\midrule
NMT~~en\biarrow cs & 96.78 & 99.27 & 98.03 \\
NMT~~en\biarrow de & 91.06 & 99.18 & 95.12 \\
NMT~~en\biarrow ru & 98.39 & 94.60 & 96.50 \\
NMT~~en\biarrow zh & 86.33 & 98.62 & 92.47 \\
\addlinespace
Macro-Avg. & 93.14 & 97.92 & 95.53 \\
\bottomrule
\end{tabularx}
\caption{Document-level classification: Accuracy of M2M-100 when detecting the translation direction of NMT translations at the document level~(documents with $\geq$ 10 sentences).}
\label{tab:doc_nmt}
\end{table}

Furthermore, Table~\ref{tab:quali_analysis} shows that our approach more easily identifies translations that are simpler in terms of verbosity and sentence complexity, e.g.: Examples 2, 3, 4. While previous research indicates that verbosity is not the most reliable feature for supervised approaches~\cite{Sominsky2019}, it is likely a helpful feature for the unsupervised approach.

Misclassified short examples as in Example 5 are not a rarity in our experiments. Our findings show that an average accuracy comparable to that reported in Table~\ref{tab:full_results_ht} is attained starting at a sentence length between 60 and 70 characters.\footnote{We used SMaLL-100 for this analysis. See Appendix~\ref{app_sentence_length}.} Additionally, we observed a trend where the accuracy of direction detection increases as the length of the sentences/documents grows. This aligns with previous unsupervised approaches, which also documented higher accuracy the larger the text chunks that were used, although there, reliable results were reported on a more extreme scale starting from text chunks with a length of 250 tokens \cite{Rabinovich2015}.

\subsection{Document-Level Classification}
\label{subsec:dl}

\noindent{}Accuracy scores for document-level results by M2M-100 (best-performing system at sentence level) are presented in Tables \ref{tab:doc_ht} (HT) and \ref{tab:doc_nmt} (NMT).
We consider documents with at least 10 sentences, and language pairs with at least 100 such documents in both directions.

The table shows that the sentence-level results are amplified at the document level. Translation direction detection accuracy for human translations reaches a macro-average of 80.5\%, while the document-level accuracy for translations generated by NMT systems reaches 95.5\% on average.

\subsection{Application to Real-World Forensic Case}
\label{subsec:rw}

Finally, we apply our approach to the 86 segment pairs of the plagiarism allegation case.
We treat the segments as a single document and classify them with M2M-100 using the document-level approach defined in Section~\ref{subsec:doc_level}.
We find that, according to the model, it is more probable that the English segments are translations of the German segments than vice versa.

We validate our analysis using a permutation test.
The null hypothesis is that the model probabilities for both potential translation directions are drawn from the same distribution.
In order to perform the permutation test, we swap the segment-level probabilities $P(y_i|x_i)$ and $P(x_i|y_i)$ for randomly selected segments $i$ before calculating the difference between the document-level probabilities $P(y|x)$ and $P(x|y)$.
We repeat this process 10,000 times and calculate the $p$-value as twice the proportion of permutations that yield a difference at least as extreme as the observed difference.
Obtaining a $p$-value of 0.0002, we reject the null hypothesis and conclude that our approach makes a statistically significant prediction that the English segments are translated from the German segments.


Overall, our analysis supports the hypothesis that German is indeed the language of origin in this real-world dataset (\textit{forgery hypothesis}; Figure~\ref{fig:figure2}). 
Nevertheless, we recommend that additional evidence of different approaches (automated, manual, and qualitative) should be considered before drawing a final conclusion, given the error rate of 5–21\% that we observed in experiments on WMT (en\biarrow de).

\section{Conclusion}
\label{sec:conclusion}
We proposed a novel approach to detecting the translation direction of parallel texts, using only an off-the-shelf multilingual NMT system.
Experiments on WMT data showed that our approach, without any task-specific supervision, is able to detect the translation direction of NMT-produced translations with relatively high accuracy, proving competitive to supervised baselines.
Accuracy increases to 96\% if the classifier is provided with at least 10 sentences per document.
We also found a robust accuracy for translations by human translators.
Finally, we applied our approach to a real-world forensic case and found that it supports the hypothesis that the English book is a forgery.
Future work should explore whether our approach can be improved by mitigating directional bias of the NMT model used.
Another open question is to what degree our approach will generalize to document-level translation.

\section*{Limitations}
While the proposed approach is simple and effective, there are some limitations that might make its application more difficult in practice:

\paragraph{Sentence alignment:}
We performed our experiments on sentence-aligned parallel data, where each sentence in one language has a corresponding sentence in the other language.
In practice, parallel documents might have one-to-many or many-to-many alignments, which would require custom pre-processing or the use of models that can directly estimate document-level probabilities.

\paragraph{Translation production workflow:}
Our main experiments used academic data from the WMT translation task, where care is taken to ensure that different translation methods are clearly separated: NMT translations did not undergo human post-editing, and human translators were instructed to work from scratch.
In practice, parallel documents might have undergone a mixture of translation strategies, which makes it more difficult to predict the accuracy of our approach.
Specifically, we found that our approach has less-than-chance accuracy on pre-NMT translations.
Applying our approach to web-scale parallel corpus filtering~\cite{Kreutzer, ranathunga-etal-2024-quality} might therefore require additional filtering steps to exclude translations of lower quality.

\paragraph{Low-resource languages:}
Our experiments required test data for both translation directions, which limited the set of languages we could test.
While the community has created reference translations for many low-resource languages, the translation directions are usually not covered symmetrically.
For example, the test set of FLORES~\cite{goyal-etal-2022-flores} has been translated from English into many languages, but not vice versa.
Thus, apart from Table~\ref{tab:lr_results_ht}, we have not tested our approach on low-resource languages, and it is possible that the accuracy of our approach is lower for such languages, in parallel with the lower translation quality of NMT models for low-resource languages.

\section*{Ethical Considerations}
Translation direction detection has a potential application in forensic linguistics, where reliable accuracy is crucial.
Our experiments show that accuracy can vary depending on the language pair, the NMT model used for detection, as well as the translation strategy and the length of the input text.
Before our approach is applied in a forensic setting, we recommend that its accuracy be validated in the context of the specific use case.

In Section~\ref{subsec:rw}, we tested our approach on a real-world instance of such a case, where one party has been accused of plagiarism, but the purported original is now suspected to be a forgery.
This case is publicly documented and has been widely discussed in German-speaking media~(e.g.,~\citealt{ebbinghaus2022b, zenthoefer2022b, dewiki:238411824}).
For this experiment, we used 86 sentence pairs from the two (publicly available) books that are the subject of this case.
However, the case has not been definitively resolved, as legal proceedings are still ongoing.
No author of this paper is involved in the legal proceedings.
We therefore refrain from publicly releasing the dataset of sentence pairs we used for this experiment.

\section*{Acknowledgements}
We thank the anonymous reviewers for their helpful comments and suggestions.
This project was funded by the Swiss National Science Foundation (project MUTAMUR; no.~213976 and project InvestigaDiff; no.~10000503).

\bibliography{bibliography}

\appendix

\onecolumn

\section{Comparison of Models (Sentence Level)}
\label{appA}

\begin{table}[h!]
\centering
\begin{tabularx}{\textwidth}{@{}Xrrrrrrrrr@{}}
\toprule
& \multicolumn{3}{c}{M2M-100-418M} & \multicolumn{3}{c}{SMaLL-100} & \multicolumn{3}{c}{NLLB-200-1.3B} \\
\cmidrule(lr){2-4} \cmidrule(lr){5-7} \cmidrule(lr){8-10}
Language Pair & \(\rightarrow\) & \(\leftarrow\) & Avg. & \(\rightarrow\) & \(\leftarrow\) & Avg. & \(\rightarrow\) & \(\leftarrow\) & Avg. \\
\midrule
NMT~~en\biarrow cs & 71.87 & 78.30 & \textbf{75.09} & 67.96 & 79.11 & 73.53 & 62.42 & 77.89 & 70.16 \\
NMT~~en\biarrow de & 62.69 & 85.27 & 73.98 & 67.05 & 80.96 & \textbf{74.00} & 73.51 & 74.36 & 73.93 \\
NMT~~en\biarrow ru & 76.91 & 71.98 & \textbf{74.44} & 74.34 & 72.30 & 73.32 & 78.87 & 57.15 & 68.01 \\
NMT~~en\biarrow uk & 75.01 & 79.31 & \textbf{77.16} & 73.74 & 78.27 & 76.01 & 58.48 & 79.56 & 69.02 \\
NMT~~en\biarrow zh & 64.29 & 90.29 & \textbf{77.29} & 66.54 & 87.62 & 77.08 & 25.42 & 90.54 & 57.98 \\
NMT~~cs\biarrow uk & 72.83 & 79.15 & 75.99 & 77.33 & 76.04 & \textbf{76.68} & 70.55 & 76.59 & 73.57 \\
NMT~~de\biarrow fr & 90.65 & 51.60 & 71.13 & 86.83 & 59.19 & \textbf{73.01} & 79.44 & 57.58 & 68.51 \\
\addlinespace
Macro-Avg. & 73.46 & 76.56 & \textbf{75.01} & 73.40 & 76.21 & 74.81 & 64.10 & 73.38 & 68.74 \\
\bottomrule
\end{tabularx}
\caption{Accuracy of three different models when detecting the translation direction of NMT-produced translations.
The first column reports accuracy for sentence pairs with left-to-right gold direction~(e.g., en\(\rightarrow\)cs), the second column for sentence pairs with the reverse gold direction~(e.g., en\(\leftarrow\)cs). The last column reports the macro-average across both directions.
The best result for each language pair is printed in bold.
}
\label{tab:full_results_nmt}
\end{table}

\vspace{0.5cm}

\begin{table}[h!]
\centering
\begin{tabularx}{\textwidth}{@{}Xrrrrrrrrr@{}}
\toprule
& \multicolumn{3}{c}{M2M-100-418M} & \multicolumn{3}{c}{SMaLL-100} & \multicolumn{3}{c}{NLLB-200-1.3B} \\
\cmidrule(lr){2-4} \cmidrule(lr){5-7} \cmidrule(lr){8-10}
Language Pair & \(\rightarrow\) & \(\leftarrow\) & Avg. & \(\rightarrow\) & \(\leftarrow\) & Avg. & \(\rightarrow\) & \(\leftarrow\) & Avg. \\
\midrule
Pre-NMT~~en\biarrow cs & 41.97 & 42.59 & \textbf{42.28} & 37.88 & 45.42 & 41.65 & 16.36 & 35.34 & 25.85 \\
Pre-NMT~~en\biarrow de & 33.33 & 54.30 & \textbf{43.81} & 36.18 & 48.57 & 42.37 & 18.73 & 26.20 & 22.47 \\
Pre-NMT~~en\biarrow ru & 37.98 & 39.01 & \textbf{38.49} & 35.71 & 39.19 & 37.45 & 19.71 & 16.04 & 17.88 \\
\addlinespace
Macro-Avg. & 37.76 & 45.30 & \textbf{41.53} & 36.59 & 44.39 & 40.49 & 18.27 & 25.86 & 22.07 \\
\bottomrule
\end{tabularx}
\caption{Accuracy of three different models when detecting the translation direction of sentences translated with \mbox{pre-NMT} systems.
The best result for each language pair is printed in bold.
}
\label{tab:full_results_prenmt}
\end{table}

\clearpage
\section{Comparison of Models (Document Level)}
\label{appB}

\begin{table}[h!]
\centering
\begin{tabularx}{\textwidth}{@{}Xrrrrrrrrr@{}}
\toprule
& \multicolumn{3}{c}{M2M-100-418M} & \multicolumn{3}{c}{SMaLL-100} & \multicolumn{3}{c}{NLLB-200-1.3B} \\
\cmidrule(lr){2-4} \cmidrule(lr){5-7} \cmidrule(lr){8-10}
Language Pair & \(\rightarrow\) & \(\leftarrow\) & Avg. & \(\rightarrow\) & \(\leftarrow\) & Avg. & \(\rightarrow\) & \(\leftarrow\) & Avg. \\
\midrule
HT~~en\biarrow cs & 88.24 & 80.62 & \textbf{84.43} & 78.92 & 89.15 & 84.03 & 55.88 & 86.05 & 70.96 \\
HT~~en\biarrow de & 70.40 & 88.43 & \textbf{79.41} & 73.60 & 82.64 & 78.12 & 68.00 & 45.45 & 56.73 \\
HT~~en\biarrow ru & 96.55 & 54.41 & 75.48 & 95.40 & 61.03 & \textbf{78.22} & 82.18 & 39.71 & 60.94 \\
HT~~en\biarrow zh & 67.65 & 97.53 & 82.59 & 71.57 & 96.30 & \textbf{83.93} & 3.92 & 96.91 & 50.42 \\
\addlinespace
Macro-Avg. & 80.71 & 80.25 & 80.48 & 79.87 & 82.28 & \textbf{81.08} & 52.50 & 67.03 & 59.76 \\
\bottomrule
\end{tabularx}
\caption{Accuracy of three different models when detecting the translation direction of human-translated documents.
The best result for each language pair is printed in bold.
}
\label{tab:full_results_doc_ht}
\end{table}

\vspace{0.5cm}

\begin{table}[h!]
\centering
\begin{tabularx}{\textwidth}{@{}Xrrrrrrrrr@{}}
\toprule
& \multicolumn{3}{c}{M2M-100-418M} & \multicolumn{3}{c}{SMaLL-100} & \multicolumn{3}{c}{NLLB-200-1.3B} \\
\cmidrule(lr){2-4} \cmidrule(lr){5-7} \cmidrule(lr){8-10}
Language Pair & \(\rightarrow\) & \(\leftarrow\) & Avg. & \(\rightarrow\) & \(\leftarrow\) & Avg. & \(\rightarrow\) & \(\leftarrow\) & Avg. \\
\midrule
NMT~~en\biarrow cs & 96.78 & 99.27 & \textbf{98.03} & 94.64 & 97.81 & 96.22 & 86.06 & 95.62 & 90.84 \\
NMT~~en\biarrow de & 91.06 & 99.18 & 95.12 & 93.85 & 97.12 & 95.49 & 96.65 & 95.06 & \textbf{95.85} \\
NMT~~en\biarrow ru & 98.39 & 94.60 & \textbf{96.50} & 97.05 & 95.12 & 96.08 & 99.20 & 72.75 & 85.97 \\
NMT~~en\biarrow zh & 86.33 & 98.62 & 92.47 & 90.62 & 98.16 & \textbf{94.39} & 13.14 & 98.39 & 55.76 \\
\addlinespace
Macro-Avg. & 93.14 & 97.92 & 95.53 & 94.04 & 97.05 & \textbf{95.55} & 73.76 & 90.46 & 82.11 \\
\bottomrule
\end{tabularx}
\caption{Accuracy of three different models when detecting the translation direction of documents translated with NMT systems.
The best result for each language pair is printed in bold.
}
\label{tab:full_results_doc_nmt}
\end{table}

\vfill





\clearpage
\section{Supervised Baseline (WMT) Result Validation Set}
\label{app_supervised}
\begin{table*}[h!]
\smaller[3]
\centering
\begin{tabularx}{\textwidth}{@{}Xrrrrrrrrrr@{}}
\toprule
& \multicolumn{3}{c}{HT} & \multicolumn{3}{c}{NMT} & \multicolumn{3}{c}{Pre-NMT} & \\
\cmidrule(lr){2-4}\cmidrule(lr){5-7}\cmidrule(lr){8-10}
Language Pair & \(\rightarrow\) & \(\leftarrow\) & Avg. & \(\rightarrow\) & \(\leftarrow\) & Avg. & \(\rightarrow\) & \(\leftarrow\) & Avg. & Overall Avg. \\
\midrule
\multicolumn{10}{l}{Learning Rate: 1e-05, Epoch 2}\\
en\biarrow cs & 83.00 & 89.00 & 86.00 & 87.00 & 89.00 & 88.00 & 89.00 & 89.00 & 89.00 & \textbf{87.67}\\
en\biarrow de & 84.00 & 78.00 & 81.00 & 85.00 & 78.00 & 81.50 & 77.00 & 82.00 & 79.50 & 80.67 \\
en\biarrow ru & 80.00 & 96.00 & 88.00 & 90.00 & 96.00 & 93.00 & 94.00 & 84.00 & 89.00 & \textbf{90.00} \\
\addlinespace
Macro-Avg. & 82.33 & 87.67 & 85.00 & 87.33 & 87.67 & 87.50 & 86.67 & 85.00 & 85.83 & 86.11\\
\midrule
\multicolumn{10}{l}{Learning Rate: 1e-05, Epoch 4}\\
en\biarrow cs & 84.00 & 88.00 & 86.00 & 90.00 & 86.00 & 88.00 & 85.00 & 92.00 & 88.50 & 87.50 \\
en\biarrow de & 92.00 & 73.00 & 82.50 & 93.00 & 74.00 & 83.50 & 71.00 & 93.00 & 82.00 & \textbf{82.67} \\
en\biarrow ru & 87.00 & 87.00 & 87.00 & 94.00 & 84.00 & 89.00 & 87.00 & 92.00 & 89.50 & 88.50 \\
\addlinespace
Macro-Avg. & 87.67 & 82.67 & 85.17 & 92.33 & 81.33 & 86.83 & 81.00 & 92.33 & 86.67 & \textbf{86.22}\\
\midrule
\multicolumn{10}{l}{Learning Rate: 1e-05, Epoch 5}\\
en\biarrow cs & 85.00 & 87.00 & 86.00 & 90.00 & 85.00 & 87.50 & 84.00 & 92.00 & 88.00 & 87.17 \\
en\biarrow de & 85.00 & 79.00 & 82.00 & 89.00 & 77.00 & 83.00 & 77.00 & 86.00 & 81.50 & 82.17\\
en\biarrow ru & 75.00 & 97.00 & 86.00 & 86.00 & 95.00 & 90.50 & 94.00 & 79.00 & 86.50 & 87.67 \\
\addlinespace
Macro-Avg. & 81.67 & 87.67 & 84.67 & 88.33 & 85.67 & 87.00 & 85.00 & 85.67 & 85.33 & 85.67\\
\midrule\midrule
\multicolumn{10}{l}{Learning Rate: 2e-05, Epoch 2}\\
en\biarrow cs & 81.00 & 88.00 & 84.50 & 82.00 & 87.00 & 84.50 & 87.00 & 84.00 & 85.50 & 84.83 \\
en\biarrow de & 67.00 & 88.00 & 77.50 & 68.00 & 90.00 & 79.00 & 89.00 & 67.00 & 78.00 & 78.17 \\
en\biarrow ru & 66.00 & 96.00 & 81.00 & 74.00 & 97.00 & 85.50 & 96.00 & 71.00 & 83.50 & 83.33\\
\addlinespace
Macro-Avg. & 71.33 & 90.67 & 81.00 & 74.67 & 91.33 & 83.00 & 90.67 & 74.00 & 82.33 & 82.11 \\
\midrule
\multicolumn{10}{l}{Learning Rate: 2e-05, Epoch 4}\\
en\biarrow cs & 82.00 & 85.00 & 83.50 & 84.00 & 87.00 & 85.50 & 83.00 & 85.00 & 84.00 & 84.33\\
en\biarrow de & 86.00 & 71.00 & 78.50 & 89.00 & 74.00 & 81.50 & 72.00 & 85.00 & 78.50 & 79.50 \\
en\biarrow ru & 85.00 & 84.00 & 84.50 & 87.00 & 84.00 & 85.50 & 85.00 & 83.00 & 84.00 & 84.67 \\
\addlinespace
Macro-Avg. & 84.33 & 80.00 & 82.17 & 86.67 & 81.67 & 84.17 & 80.00 & 84.33 & 82.17 & 82.83\\
\midrule
\multicolumn{10}{l}{Learning Rate: 2e-05, Epoch 5}\\
en\biarrow cs & 91.00 & 84.00 & 87.50 & 89.00 & 80.00 & 84.50 & 82.00 & 91.00 & 86.50 & 86.17 \\
en\biarrow de & 83.00 & 76.00 & 79.50 & 87.00 & 78.00 & 82.50 & 78.00 & 84.00 & 81.00 & 81.00 \\
en\biarrow ru & 83.00 & 85.00 & 84.00 & 85.00 & 88.00 & 86.50 & 86.00 & 85.00 & 85.50 & 85.33 \\
\addlinespace
Macro-Avg. & 85.67 & 81.67 & 83.67 & 87.00 & 82.00 & 84.50 & 82.00 & 86.67 & 84.33 & 84.17\\
\midrule\midrule
\multicolumn{10}{l}{Learning Rate: 3e-05, Epoch 2}\\
en\biarrow cs & 81.00 & 91.00 & 86.00 & 81.00 & 89.00 & 85.00 & 89.00 & 82.00 & 85.50 & 85.50\\
en\biarrow de & 0.00 & 100.00 & 50.00 & 0.00 & 100.00 & 50.00 & 100.00 & 0.00 & 50.00 & 50.00\\
en\biarrow ru & 59.00 & 98.00 & 78.50 & 61.00 & 99.00 & 80.00 & 98.00 & 55.00 & 76.50 & 78.33\\
\addlinespace
Macro-Avg. & 46.67 & 96.33 & 71.50 & 47.33 & 96.00 & 71.67 & 95.67 & 45.67 & 70.67 & 66.83\\
\midrule
\multicolumn{10}{l}{Learning Rate: 3e-05, Epoch 4}\\
en\biarrow cs & 77.00 & 92.00 & 84.50 & 80.00 & 91.00 & 85.50 & 88.00 & 80.00 & 84.00 & 84.67\\
en\biarrow de & 100.00 & 0.00 & 50.00 & 100.00 & 0.00 & 50.00 & 0.00 & 100.00 & 50.00 & 50.00\\
en\biarrow ru & 63.00 & 98.00 & 80.50 & 72.00 & 98.00 & 85.00 & 98.00 & 65.00 & 81.50 & 82.33\\
\addlinespace
Macro-Avg. & 80.00 & 63.33 & 71.67 & 84.00 & 63.00 & 73.50 & 62.00 & 81.67 & 71.83 & 72.33\\
\midrule
\multicolumn{10}{l}{Learning Rate: 3e-05, Epoch 5}\\
en\biarrow cs & 75.00 & 91.00 & 83.00 & 76.00 & 88.00 & 82.00 & 90.00 & 75.00 & 82.50 & 82.50\\
en\biarrow de & 100.00 & 0.00 & 50.00 & 100.00 & 0.00 & 50.00 & 0.00 & 100.00 & 50.00 & 38.89\\
en\biarrow ru & 69.00 & 98.00 & 83.50 & 75.00 & 96.00 & 85.50 & 95.00 & 71.00 & 83.00 & 84.00\\
\addlinespace
Macro-Avg. & 81.33 & 63.00 & 72.17 & 83.67 & 61.33 & 72.50 & 61.67 & 82.00 & 71.83 & 72.17\\
\midrule\midrule
\multicolumn{10}{l}{Learning Rate: Dynamic, Epoch 2}\\
en\biarrow cs & 83.00 & 88.00 & 85.50 & 80.00 & 85.00 & 82.50 & 88.00 & 82.00 & 85.00 & 84.33\\
en\biarrow de & 81.00 & 83.00 & 82.00 & 79.00 & 83.00 & 81.00 & 85.00 & 79.00 & 82.00 & 81.67\\
en\biarrow ru & 83.00 & 90.00 & 86.50 & 89.00 & 90.00 & 89.50 & 90.00 & 82.00 & 86.00 & 87.33\\
\addlinespace
Macro-Avg. & 82.33 & 87.00 & 84.67 & 82.67 & 86.00 & 84.33 & 87.67 & 81.00 & 84.33 & 84.44\\
\midrule
\multicolumn{10}{l}{Learning Rate: Dynamic, Epoch 4}\\
en\biarrow cs & 83.00 & 89.00 & 86.00 & 85.00 & 90.00 & 87.50 & 88.00 & 83.00 & 85.50 & 86.33\\
en\biarrow de & 88.00 & 75.00 & 81.50 & 93.00 & 78.00 & 85.50 & 78.00 & 83.00 & 80.50 & 82.50\\
en\biarrow ru & 79.00 & 88.00 & 83.50 & 87.00 & 88.00 & 87.50 & 90.00 & 81.00 & 85.50 & 85.50\\
\addlinespace
Macro-Avg. & 83.33 & 84.00 & 83.67 & 88.33 & 85.33 & 86.83 & 85.33 & 82.33 & 83.83 & 84.78\\
\midrule
\multicolumn{10}{l}{Learning Rate: Dynamic, Epoch 5}\\
en\biarrow cs & 75.00 & 92.00 & 83.50 & 77.00 & 90.00 & 83.50 & 90.00 & 80.00 & 85.00 & 84.00\\
en\biarrow de & 88.00 & 77.00 & 82.50 & 88.00 & 78.00 & 83.00 & 77.00 & 82.00 & 79.50 & 81.67\\
en\biarrow ru & 79.00 & 91.00 & 85.00 & 87.00 & 88.00 & 87.50 & 89.00 & 79.00 & 84.00 & 85.50\\
\addlinespace
Macro-Avg. & 80.67 & 86.67 & 83.67 & 84.00 & 85.33 & 84.67 & 85.33 & 80.33 & 82.83 & 83.72 \\
\midrule\midrule
\end{tabularx}
\caption{Accuracy of the supervised baseline when detecting the translation direction of HT, NMT, and Pre-NMT-produced translations across various checkpoints and learning rates on the WMT validation set.}
\label{tab:combined_results}
\end{table*}

\clearpage
\section{Supervised Baseline (WMT) Result Validation Set with Joint Encoding}
\label{app:joint_enc}

\begin{table*}[h!]
\centering
\begin{tabularx}{\textwidth}{@{}lXrrrrrrr@{}}
\toprule
LP& Input Order& \multicolumn{3}{c}{HT} & \multicolumn{3}{c}{NMT} \\
\cmidrule(lr){3-5} \cmidrule(lr){6-8}
 & & \(\rightarrow\) & \(\leftarrow\) & Avg. & \(\rightarrow\) & \(\leftarrow\) & Avg. \\
\midrule
\multirow{3}{*}{en\biarrow de}
& source + translation & 79.51 & 65.45 & 72.48 & 83.13 & 69.23 & 76.18 \\
& translation + source & 78.59 & 60.86 & 69.72 & 83.91 & 65.57 & 74.74 \\
\addlinespace
\multirow{3}{*}{en\biarrow ru}
& source + translation & 66.13 & 80.73 & 73.43 & 70.31 & 81.88 & 76.1 \\
& translation + source & 67.64 & 83.24 & 75.44 & 69.57 & 84.05 & 76.81 \\
\addlinespace
\multirow{3}{*}{en\biarrow cs}
& source + translation & 61.26 & 69.3 & 65.28 & 62.24 & 69.65 & 65.94 \\
& translation + source & 61.7 & 66.89 & 64.3 & 63.15 & 67.51 & 65.33\\
\bottomrule
\end{tabularx}
\caption{WMT test set results of the joint encoding approach, where the translated segment is appended to the source and vice versa.}
\label{tab:comparison_results}
\end{table*}

\subsection*{Model and Data}
In addition to the siamese architecture outlined in Section~\ref{subsec:supervised}, which encodes each text segment independently before concatenating their respective embeddings, we also investigate an alternative approach that involves joint encoding of the source and translation texts. Specifically, the two text segments are concatenated with a special delimiter token (``</s>'') inserted between them and then passed together through the encoder \cite{ranasinghe-etal-2020-transquest, rei-etal-2022-cometkiwi}. The joint encoding system is trained using the same hyperparameters as outlined in Section~\ref{subsec:supervised}, ensuring a consistent comparison.

To mitigate potential input order bias, we reversed the input order for half of the training data. 
However, a comparison of the results for the WMT test set with differing input order of the source and the target (Table~\ref{tab:comparison_results}) shows that this strategy did not eliminate the bias.

Ultimately, we observed that the siamese architecture demonstrated greater stability with respect to input order, without incurring a significant performance trade-off. Based on these findings, we focus on the siamese architecture as the main baseline presented in this paper.



\clearpage
\section{Supervised Baseline (Europarl) and Cross-Domain Application }
\label{app_europarl}

\begin{table*}[h!]
\centering
\begin{tabularx}{\textwidth}{@{}lXrrrrrrrrr@{}}
\toprule
LP & Approach & \multicolumn{3}{c}{HT (Europarl)} & \multicolumn{3}{c}{HT (WMT)} & \multicolumn{3}{c}{NMT (WMT)} \\
\cmidrule(lr){3-5} \cmidrule(lr){6-8} \cmidrule(lr){9-11}
& & \(\rightarrow\) & \(\leftarrow\) & Avg. & \(\rightarrow\) & \(\leftarrow\) & Avg. & \(\rightarrow\) & \(\leftarrow\) & Avg. \\
\midrule
\multirow{3}{*}{en\biarrow cs} 
& sup (WMT) & --- & --- & --- & 64.70 & \textbf{71.36} & \textbf{68.03} & 65.43 & 71.98 & 68.71 \\
& sup (Europarl) & 71.51 & 83.14 & 77.32 & \textbf{82.03} & 48.39 & 65.21 & \textbf{82.72} & 50.18 & 66.45 \\
& unsup & --- & --- & --- & 68.85 & 65.19 & 67.02 & 71.87 & \textbf{78.30} & \textbf{75.09} \\
\addlinespace
\multirow{3}{*}{en\biarrow de}
& sup (WMT) & --- & --- & --- & \textbf{87.05} & 55.29 & \textbf{71.17} & \textbf{89.52} & 61.08 & \textbf{75.30} \\
& sup (Europarl)& 82.12 & 70.80 & 76.46 & 80.97 & 45.36 & 63.17 & 84.15 & 39.70 & 61.93 \\
& unsup & --- & --- & --- & 56.38 & \textbf{67.44} & 61.91 & 62.69 & \textbf{85.27} & 73.98 \\
\addlinespace
\multirow{3}{*}{en\biarrow ru}
& sup (WMT) & --- & --- & --- & 56.65 & \textbf{81.74} & \textbf{69.19} & 61.72 & \textbf{82.97} & 72.35 \\
& sup (Europarl) & --- & --- & --- & --- & --- & --- & --- & --- & --- \\
& unsup & --- & --- & --- & \textbf{71.81} & 54.05 & 62.93 & \textbf{76.91} & 71.98 & \textbf{74.44} \\
\addlinespace
\multirow{3}{*}{de\biarrow fr}
& sup (WMT) & --- & --- & --- & --- & --- & --- & --- & --- & --- \\
& sup (Europarl) & 59.80 & 56.67 & 58.24 & 58.28 & 52.78 & 55.53 & 58.85 & 53.38 & 56.11 \\
& unsup & --- & --- & --- & --- & --- & --- & --- & --- & --- \\
\bottomrule
\end{tabularx}
\caption{Comparison of supervised and unsupervised approaches on the translation direction detection task for HT and NMT datasets from WMT and Europarl. The supervised approach involves fine-tuning XLM-R on Europarl as well as on WMT and testing on WMT.}
\label{tab:europarl1}
\end{table*}




\begin{table}[h!]
\centering
\begin{tabularx}{\textwidth}{@{}Xrrr@{}}
\toprule
Language Pair & HT (Europarl)& HT (WMT) &NMT (WMT)\\
\midrule
de\biarrow fr & 0.031 & 0.055 & 0.055  \\
cs\biarrow en & 0.116 & 0.336 & 0.325  \\
de\biarrow en & 0.113 & 0.356 & 0.444  \\
\addlinespace
Macro-Avg. & 0.087 & 0.249 & 0.273  \\
\bottomrule
\end{tabularx}
\caption{Bias values across Europarl and WMT test sets.}
\label{tab:europarl_bias}
\end{table}
\vspace{1cm}

\subsection*{Model and Data}
\noindent In addition to the supervised system trained on WMT16 data, we train a supervised system on a subset of Europarl~\cite{Koehn2005, Ustaszewski}, a corpus consisting of parallel text from the proceedings of the European Parliament. We train a system for each language pair de\biarrow fr, cs\biarrow en, de\biarrow en as described in Subsection~\ref{subsec:supervised}. We train on 10,249 samples per direction, use a validation and test set of 1,281 further samples each per direction. Additionally, we apply the systems to the corresponding WMT test data subset as in Subsection~\ref{subsec:supervised}, but with WMT22 de\biarrow fr instead of WMT22/23 en\biarrow ru, to test the systems' cross-domain capabilities. We experiment in the same manner with several learning rates and epochs as in Appendix~\ref{app_supervised} and select the best performing system for each language pair based on the validation accuracy (de\biarrow fr: lr = dynamic, epoch = 5; cs\biarrow en: lr = dynamic, epoch = 5; de\biarrow en: lr = 1e-05, epoch = 5). 

\subsection*{Results}
Table~\ref{tab:europarl1} shows the results for the unsupervised approach, as well as both the WMT- and Europarl-based supervised systems. While the systems reach higher in-domain accuracies for en\biarrow cs and en\biarrow de than the WMT-based system, for de\biarrow fr the scores are substantially lower. The results decrease further when the system is tested on out-of-domain data, namely the WMT newstest data. On average, a decrease of 9.37\% for WMT human translations and 9.18\% NMT-based WMT translations is observable. These results on cross-domain applicability for supervised translation direction detection align with previous work~\cite{Sominsky2019}.
A further notable observation, when comparing the in-domain to out-of-domain results, is the increased directional bias, which triples when the systems are applied to out-of-domain data (Table~\ref{tab:europarl_bias}).

\clearpage

\section{Comparison of Results for Individual WMT Datasets}
\label{app_contamination}

\begin{table*}[h!]
\smaller[1]
\centering
\begin{tabularx}{\textwidth}{@{}Xrrrrrrrrr@{}}
\toprule
& \multicolumn{3}{c}{M2M-100-418M} & \multicolumn{3}{c}{SMaLL-100} & \multicolumn{3}{c}{NLLB-200-1.3B} \\
\cmidrule(lr){2-4} \cmidrule(lr){5-7} \cmidrule(lr){8-10}
Language Pair & WMT16 & WMT22 & WMT23 & WMT16 & WMT22 & WMT22 & WMT16 & WMT22 & WMT23 \\
\midrule
en\biarrow cs & 66.43 & 73.44 & 71.77 & 64.44 & 71.76 & 67.57 & 50.72 & 67.09 & 62.77 \\
en\biarrow de & 61.88 & 71.65 & - & 63.34 & 71.53 & - & 46.53 & 70.27 & - \\
en\biarrow ru & 63.91 & 73.34 & 74.05 & 62.29 & 72.28 & 72.83 & 44.30 & 65.20 & 70.60 \\
en\biarrow uk & - & 78.98 & 74.38 & - & 77.34 & 73.68 & - & 67.53 & 67.85 \\
en\biarrow zh & - & 76.18 & 75.60 & - & 75.97 & 75.52 & - & 51.97 & 58.87 \\
cs\biarrow uk & - & 75.39 & 72.09 & - & 75.47 & 76.63 & - & 72.89 & 70.08 \\
de\biarrow fr & - & 69.58 & - & - & 71.82 & - & - & 68.48 & - \\
\addlinespace
Macro-Avg. & 64.08 & 74.08 & 73.58 & 63.36 & 73.74 & 73.25 & 47.19 & 66.20 & 66.03 \\
\bottomrule
\end{tabularx}
\caption{Average accuracy for language pairs for the individual WMT datasets and tested models. 
}
\label{tab:wmt_accuracy}
\end{table*}

\subsection*{Results by WMT year and model}

The WMT16 test data predates the release of all models in our experiments. Additionally, the WMT22 data was released before the NLLB model. This circumstance raises the possibility of data contamination, as the models may have been exposed to test data during training. To address this concern, we compare results across different WMT datasets and add test samples from WMT23, which was released after all three models. If the accuracies for a dataset predating the release of a model are noticeably higher than for a more recent dataset, it would suggest data contamination. However, as shown in~\ref{tab:wmt_accuracy}, this is not observed, indicating that our results are likely unaffected by data contamination.

\clearpage

\vfill

\clearpage

\section{Data Statistics}
\label{sec:appendix_data}

\begin{table}[h!]
\centering
\begin{tabular}{llrrrrr}
\toprule
& & \multicolumn{2}{c}{Source} & \multicolumn{3}{c}{Reference} \\
Test Set & Direction & Sents & Docs $\geq10$ & HT & NMT & Pre-NMT \\
\cmidrule(r){3-4} \cmidrule(l){5-7}
WMT16 & cs$\rightarrow$en & 1499 & 40 & \textit{1499} & \textit{1499} & 16489 \\
WMT16 & de$\rightarrow$en & 1499 & 55 & \textit{1499} & \textit{1499} & 13491 \\
WMT16 & en$\rightarrow$cs & 1500 & 54 & \textit{1500} & \textit{3000} & 27000 \\
WMT16 & en$\rightarrow$de & 1500 & 54 & \textit{1500} & \textit{4500} & 18000 \\
WMT16 & en$\rightarrow$ru & 1500 & 54 & \textit{1500} & \textit{3000} & 15000 \\
WMT16 & ru$\rightarrow$en & 1498 & 52 & \textit{1498} & \textit{1498} & 13482 \\
\midrule
WMT22 & cs$\rightarrow$en & 1448 & 129 & 2896 & 15928 & - \\
WMT22 & cs$\rightarrow$uk & 1930 & 13 & 1930 & 23160 & - \\
WMT22 & de$\rightarrow$en & 1984 & 121 & 3968 & 17856 & - \\
WMT22 & de$\rightarrow$fr & 1984 & 73 & 1984 & 11904 & - \\
WMT22 & en$\rightarrow$cs & 2037 & 125 & 4074 & 20370 & - \\
WMT22 & en$\rightarrow$de & 2037 & 125 & 4074 & 18333 & - \\
WMT22 & en$\rightarrow$ru & 2037 & 95 & 2037 & 22407 & - \\
WMT22 & en$\rightarrow$uk & 2037 & 95 & 2037 & 18333 & - \\
WMT22 & en$\rightarrow$zh & 2037 & 125 & 4074 & 26481 & - \\
WMT22 & fr$\rightarrow$de & 2006 & 71 & 2006 & 14042 & - \\
WMT22 & ru$\rightarrow$en & 2016 & 73 & 2016 & 20160 & - \\
WMT22 & uk$\rightarrow$cs & 2812 & 43 & 2812 & 33744 & - \\
WMT22 & uk$\rightarrow$en & 2018 & 22 & 2018 & 20180 & - \\
WMT22 & zh$\rightarrow$en & 1875 & 102 & 3750 & 22500 & - \\
\midrule
WMT23 & cs$\rightarrow$uk & 2017 & 99 & 2017 & 26221 & - \\
WMT23 & en$\rightarrow$cs & 2074 & 79 & 2074 & 31110 & - \\
WMT23 & en$\rightarrow$ru & 2074 & 79 & 2074 & 24888 & - \\
WMT23 & en$\rightarrow$uk & 2074 & 79 & 2074 & 22814 & - \\
WMT23 & en$\rightarrow$zh & 2074 & 79 & 2074 & 31110 & - \\
WMT23 & ru$\rightarrow$en & 1723 & 63 & 1723 & 20676 & - \\
WMT23 & uk$\rightarrow$en & 1826 & 66 & 1826 & 20086 & - \\
WMT23 & zh$\rightarrow$en & 1976 & 60 & 1976 & 29640 & - \\
\bottomrule
\end{tabular}
\caption{Detailed data statistics for the main experiments.  
Cursive: data used for validation.}
\label{tab:data_stats}
\end{table}

\vspace{0.5cm}


\begin{table}[h!]
\centering
\begin{tabular}{lr}
\toprule
Direction & Sentence Pairs \\
\midrule
bn$\leftrightarrow$hi & 1012 \\
de$\leftrightarrow$hi & 1012 \\
cs$\leftrightarrow$uk & 1012 \\
de$\leftrightarrow$fr & 1012 \\
zh$\leftrightarrow$fr & 1012 \\
xh$\leftrightarrow$zu & 1012 \\
\bottomrule
\end{tabular}
\caption{Statistics for the FLORES-101 (devtest) datasets, where both sides are human translations from English.}
\label{tab:flores_stats}
\end{table}

\clearpage
\section{Sentence Length}
\label{app_sentence_length}


\begin{table}[h!]
\centering
\small
\begin{tabular}{lrrrrrrrrrr}
\toprule
Language Pair & 0--19 & 20--39 & 40--59 & 60--79 & 80--99 & 100--119 & 120--139 & 140--159 & 160--179 & 180--199 \\
\midrule
NMT en-cs   & 59.41 & 62.46 & 65.88 & 69.62 & 74.42 & 75.28 & 74.93 & 78.92 & 79.08 & 81.45 \\
NMT en-de   & 63.97 & 67.20 & 66.24 & 71.99 & 76.38 & 76.21 & 79.26 & 79.34 & 80.50 & 84.60 \\
NMT en-ru   & 64.82 & 62.99 & 68.73 & 71.80 & 75.44 & 76.50 & 79.42 & 79.12 & 81.16 & 81.33 \\
NMT en-uk   & 64.59 & 67.58 & 72.44 & 75.31 & 78.63 & 79.67 & 81.06 & 78.85 & 84.22 & 79.90 \\
NMT en-zh   & 74.32 & 73.82 & 77.00 & 81.31 & 82.84 & 83.75 & 84.33 & 83.49 & 86.23 & 87.12 \\
NMT cs-uk   & 61.93 & 69.72 & 76.24 & 76.39 & 80.21 & 80.69 & 83.42 & 81.46 & 83.07 & 84.11 \\
NMT de-fr   & 57.44 & 65.25 & 70.20 & 74.27 & 76.20 & 73.70 & 79.93 & 72.60 & 76.90 & 78.63 \\
\addlinespace
Macro Avg.  & 63.78 & 67.00 & 70.96 & 74.38 & 77.73 & 77.97 & 80.34 & 79.11 & 81.59 & 82.45 \\
\bottomrule
\end{tabular}
\caption{Translation direction detection accuracy by number of characters in the source sentence for different language pairs using SMaLL-100.}
\label{tab:sent_len}
\end{table}

\bigskip

\section{Example for Forensic Dataset}
\label{app_real_world}

\begin{table*}[h!]
\centering
\begin{tabularx}{\textwidth}{@{}Xrrr@{}}
  \toprule
  & DE\(\rightarrow\)EN & EN\(\rightarrow\)DE \\
  \midrule
  \textit{DE: Nach 30 sec. wurde das trypsinhaltige PBS abgegossen, und die Zellen kamen für eine weitere Minute in den Brutschrank.} & & \\
  \addlinespace
  EN: After 30 sec, the trypsin-containing PBS was poured off, and the cells were placed in the incubator for another minute. & 0.442 & 0.285 \\
  \bottomrule
\end{tabularx}
\caption{Example of two segments from the forensic case~(Section~\ref{subsec:rw}). M2M-100 assigns a higher probability to the English sentence conditioned on the German sentence than vice versa, suggesting that the English sentence is more likely to be a translation of the German sentence.}
\label{tab:examples-colchicine}
\end{table*}

\clearpage

\end{document}